# Explaining black-box text classifiers for disease-treatment information extraction

**Milad Moradi, Matthias Samwald**

Institute for Artificial Intelligence and Decision Support
Medical University of Vienna, Austria
`{milad.moradivastegani, matthias.samwald}`@meduniwien.ac.at

## Abstract

Deep neural networks and other intricate Artificial Intelligence (AI) models have reached high levels of accuracy on many biomedical natural language processing tasks. However, their applicability in real-world use cases may be limited due to their vague inner working and decision logic. A post-hoc explanation method can approximate the behavior of a black-box AI model by extracting relationships between feature values and outcomes. In this paper, we introduce a post-hoc explanation method that utilizes confident itemsets to approximate the behavior of black-box classifiers for medical information extraction. Incorporating medical concepts and semantics into the explanation process, our explanator finds semantic relations between inputs and outputs in different parts of the decision space of a black-box classifier. The experimental results show that our explanation method can outperform perturbation and decision set based explanators in terms of fidelity and interpretability of explanations produced for predictions on a disease-treatment information extraction task.

## 1 Introduction

Intricate Artificial Intelligence (AI) methods such as deep neural networks, support vector machines, ensemble methods, etc. have achieved high levels of accuracy in many Biomedical Natural Language Processing (BioNLP) and text analysis applications (Moradi, Dashti, & Samwald, 2020; Moradi, Dorffner, & Samwald, 2020; Wu et al., 2019). However, a big challenge is that computational models learned through training these black-boxes are complex and poorly interpretable. This unintelligibility problem may be much more severe in the biomedical and healthcare domains since there is a crucial need to explain outcomes of a black-box system in a human-understandable form (Gao, Liu, Lawley, & Hu, 2017). eXplainable AI (XAI) methods address this problem by revealing useful information about the decision-making process of a black-box, or by approximating its behavior (Guidotti et al., 2018).

Explaining black-box text classifiers may be too challenging to be properly handled by common perturbation or rule-based methods (Lakkaraju, Kamar, Caruana, & Leskovec, 2019; Ribeiro, Singh, & Guestrin, 2016). First, text data are inherently symbolic; changing a text sample's numerical representation may lead to producing a meaningless sample. Second, word embeddings are not easily interpretable. Even if the importance of a dimension is determined when predicting a class, it can be impossible to interpret what that dimension means. Third, semantic perturbations need to be performed carefully. In contrast to numerical data that can be randomly perturbed with respect to the range of values that a feature can take, perturbing textual samples with respect to their semantics need careful utilization of ontologies or human supervision. Fourth, the vocabulary in a NLP task may be very large, making it difficult to explain a text classifier using a manageable number of rules.

In this paper, we address the above challenges in the context of medical text classification. We introduce a model-agnostic, post-hoc method that combines confident itemset mining and medical domain knowledge to produce explanations for black-box text classifiers applied to disease-treatment information extraction. Since the decision boundaries of a multi-label classification problem may be complex, our explanation method discretizes the whole input space into smaller subspaces to facilitate producing high-quality local explanations. Within every subspace, confident itemsets are extracted to represent those medical concepts that are highly related to a specific class label. The extracted confident itemsets are used for generating instance-wise explanations.

We evaluated the fidelity and interpretability of our explanation method on a disease-treatment information extraction task. The results show that our method can produce more accurate and interpretable explanations than the perturbation-based and decision set explanators.

## 2 Related work

XAI methods can be generally divided into two groups, i.e. intrinsic versus post-hoc explainability (Du, Liu, & Hu, 2019). Intrinsic explainability refers to those AI models that are explainable by design. These models include linear regression models, decision trees, and decision rules. On the other hand, post-hoc explanations are constructed after building a black-box model. In fact, the black-box, uninterpretable model is mapped to a proxy model that can be easily explained and understood (Keane & Kenny, 2019). Our explanation method falls in the post-hoc category, where the predictive accuracy of the target AI system remains unaffected since no explainability consideration restricts the model's complexity. The proxy model can reflect the statistical properties of the black-box as long as the explanations accurately approximate the behavior of the complex model (Du et al., 2019).

During recent years, some studies addressed XAI for BioNLP applications. Gao et al. (Gao et al., 2017) combined labeled sequential patterns, UMLS semantic types, sentence-based, and heuristic features in order to build decision rules from decision tress of a forest-based model for information extraction from online health forums. Gehrmann et al. (Gehrmann et al., 2018) utilized a phrase-saliency method to explain outcomes of a convolutional neural network for patient phenotyping from clinical narratives. In this paper, we focus on post-hoc explanation of black-box classifiers applied in disease-treatment information extraction. The model-agnosticism property of our method allows to explain any black-box regardless of the underlying model.

## 3 The explanation method

Let $f: X \rightarrow C$ be a black-box text classifier trained on a training set, $X=\{X_1, X_2, …, X_M\}$ be a dataset of $M$ textual instances, and $C=\{C_1, C_2, …, C_Q\}$ be the set of class labels in a classification problem. Given a class label $Y_m \in C$ predicted by $f$ as the prediction for $X_m$, the goal is to explain the local behavior of $f$ using a set of itemsets that refer to those medical concepts that appear in $X_m$ and are highly correlated with $Y_m$.

**Step 1: Mapping text to concepts.** Every instance $X_m$ is originally represented as a set of $P$ words such that $X_m=\{word_1, …, word_P\}$. Using the MetaMap tool, every instance $X_m$ is mapped to a set of medical concepts contained in the Unified Medical Language System (UMLS) (Lindberg, Humphreys, & McCray, 1993), such that $X'_m=\{concept_1, …, concept_B\}$ is the concept-based representation of $X_m$.

**Step 2: Extracting confident itemsets.** Since the whole decision space of a multi-label classification task may be complex and nonlinear, building an accurate global explanation model can be infeasible. In order to deal with this issue, we discretize the decision space into smaller subspaces and approximate the local behaviour of the black-box model in every subspace. The decision space $S$ is discretized into subspaces $\{S_1, …, S_Q\}$ such that subspace $S_q$ is characterized by class $C_q$ and contains those instances that were classified in $C_q$ by the black-box classifier $f$.

Frequent itemset mining is a technique to discover frequent local patterns in a dataset. However, when working with text data, frequent concepts may not be useful to discriminate between classes. In order to tackle this problem, we introduce confident itemsets that capture those concepts that are highly related to a class label, even if they are infrequent.

The confident itemset mining step begins by representing every instance $X_m$ as a set of items $\{<item_1>, …, <item_D>\}$ where every item $<item_d>$ corresponds to a concept in the concept-based representation of $X_m$. In the $K_{th}$ iteration of the confident itemset mining algorithm, a set of confident $K$-itemsets is extracted for every subspace $S_q$. Given a subspace $S_q$, a confident $K$-itemset $ci$ is a set of $K$ distinct items that satisfies two criteria: 1) the confidence property of $ci$ within the class $C_q$ must be equal to or greater than a confidence threshold *min_conf*, and 2) every subset of *ci* must be a confident itemset within the class $C_q$. The confidence property of *ci* is computed within class $C_q$ as follows:

$$Confidence(ci, C_q) = \frac{P(ci|C_q)}{P(ci)} \quad (1)$$

where $P(ci)$ is the probability of observing *ci* in dataset *X*, and $P(ci|C_q)$ is the probability of observing *ci* in instances belonging to class $C_q$. The confident itemset mining process continues until *K* reaches a predefined value or no confident itemset is extracted. A detailed description of the confident itemset mining algorithm can be found in (Moradi & Samwald, 2021).

After finishing the itemset mining algorithm, a set of confident itemsets have been extracted for every class $C_q$ that approximate the decision boundaries of the respective subspace $S_q$.

**Step 3: Instance-wise explanations.** The extracted confident itemsets show relationships between concepts and class labels in different subspaces and can be used to approximate the local behaviour of the black-box model for individual predictions. Given an instance $X_m$ and its class label $Y_m$ predicted by the black-box *f*, and a set of confident itemsets $CI_q=\{ci_1, …, ci_J\}$ extracted for every class $C_q$, an instance-wise explanation $E_m=\{<ci_1, …, ci_U>, Y'_m\}$ is generated for instance $X_m$, where $<ci_1, …, ci_U>$ is a set of confident itemsets and $Y'_m$ is the class label assigned to $X_m$ by our explanation method.

The set $<ci_1, …, ci_U>$ is constructed by searching every $CI_q$ and extracting those itemsets that appear in $X_m$. A confidence score is computed for $E_m$ for every class $C_q$ that has at least one itemset in the set $<ci_1, …, ci_U>$, as follows:

$$Confidence\_score(E_m, C_q) = \sum_{u=1}^{U} Confidence(ci_u, C_q) \quad (2)$$

where $Confidence\_score(E_m, C_q)$ is the confidence score of explanation $E_m$ in class $C_q$, and $Confidence(ci_u, C_q)$ is the confidence value of itemset $ci_u$ in class $C_q$.

Finally, the explanation method assigns a class label $Y'_m$ to instance $X_m$ by selecting the class obtaining the highest confidence score for explanation $E_m$. If two classes obtain the same score, the class label that appears more frequently is selected. If no classes obtain a confidence score more than zero, the most frequent class label in the classification problem is selected.

## 4 Experimental results

We used the BioText disease-treatment information extraction dataset (Rosario & Hearst, 2004) to evaluate the performance of our explanation method. The dataset contains 3,655 short text passages classified into eight classes based on the semantic relationships between diseases and treatments mentioned in the text. We divided the dataset into a training and a test set containing 2,924 and 731 samples, respectively. We trained two black-box text classifier models based on BioBERT (Lee et al., 2019) and a Long Short-Term Memory (LSTM) network (Hochreiter & Schmidhuber, 1997) on the training set. We tested the classifiers on the test set and produced explanations for the predictions made on the test samples. We compared our explanation method with Local Interpretable Model-agnostic Explanations (LIME) (Ribeiro et al., 2016) and Model Understanding through Subspace Explanations (MUSE) (Lakkaraju et al., 2019). LIME creates a linear model based on perturbing a sample to approximate the local behavior of a model. MUSE discovers frequent itemsets to construct a decision set containing if-then rules that represent decision logic of the target model. We implemented a variation of MUSE that uses concepts instead of numerical and categorical features in the condition part of a decision set. We set the value of the parameter *min_conf* of our CIE method to 0.7 since it led to the best trade-off between the size of explanations and fidelity.

### 4.1 Fidelity

We evaluated the explanation methods with respect to the fidelity measure that quantifies how accurately the explanator can mimic the behavior of the black-box in terms of assigning class labels to the samples. Table 1 presents the fidelity scores obtained by our Confident Itemsets Explanation (CIE) method and the other explanators on the predictions of the two black-box classifiers. Figure 1 shows explanations produced by our CIE method for three classifications made by the BioBERT classifier. CIE assigned correct class labels to the three samples, as same as the target black-box.

|       | BioBERT | LSTM  |
|-------|---------|-------|
| LIME  | 0.849   | 0.836 |
| MUSE  | 0.775   | 0.771 |
| CIE   | 0.917   | 0.908 |

Table 1: The fidelity scores obtained by the explanation methods.

| Sample | Antiplatelet therapy with aspirin and systematic anticoagulation with warfarin reduce cardiovascular morbidity and mortality after myocardial infarction when given alone. | |
|---|---|---|
| Predicted class | TREATMENT FOR DISEASE | |
| Explanation | *Itemset* | *Confidence* |
| | <Aspirin therapy> | 1.0 |
| | <Cardiovascular morbidity> | 0.79 |
| Sample | Sequelae include severe developmental delay and asymmetric double hemiplegia. | |
| Predicted class | DISORDER ONLY | |
| Explanation | *Itemset* | *Confidence* |
| | <sequelae aspects> | 1.0 |
| | <Developmental delay, severe> | 0.75 |
| | <Hemiplegia> | 0.75 |
| Sample | They underwent neurologic assessment , brain computed tomography , IQ testing , and assessment with the Childhood Behavior Checklist at baseline before methylphenidate therapy. | |
| Predicted class | TREATMENT ONLY | |
| Explanation | *Itemset* | *Confidence* |
| | <CAT scan of brain> | 1.0 |
| | <Intelligence Tests> | 1.0 |
| | <child behavior checklist> | 0.84 |
| | <Neurologic Examination> | 0.72 |

Figure 1: Explanations produced by our CIE method for three classifications from the BioText dataset.

| $N$ | LIME  | MUSE  | CIE   |
|-----|-------|-------|-------|
| 10  | 0.641 | 0.633 | 0.698 |
| 20  | 0.690 | 0.669 | 0.752 |
| 30  | 0.757 | 0.716 | 0.819 |
| 40  | 0.795 | 0.778 | 0.842 |
| 50  | 0.831 | 0.805 | 0.890 |

Table 2: The fidelity scores for different sizes of class-wise explanations on the outcomes of the BioBERT classifier.

### 4.2 Interpretability

We assessed the interpretability of the explanations in terms of the trade-off between the size of class-wise explanations and the fidelity. Given a parameter $N$ that specifies the maximum number of decision units in a class-wise explanation, we evaluated the ability of the methods in producing accurate explanations using as few decision units as possible. A decision unit refers to a confident itemset in CIE, a concept and its coefficient in the linear model in LIME, and an if-then rule in MUSE. Class-wise explanations were produced for CIE by selecting the top $N$ confident itemsets extracted from a particular class. For LIME, the top $N$ concepts with the largest coefficients in the samples classified in a specific class are selected as the class-wise explanation. For MUSE, a class-wise explanation is generated by selecting $N$ if-then rules that lead to assigning the respective class label. Table 2 and 3 present the fidelity scores obtained by the explanation methods on the predictions of the BioBERT and LSTM classifiers, respectively, when $S$=10, 20, 30, 40, and 50. It is worth to note that a subjective human evaluation already showed CIE can generate more interpretable and human-understandable explanations than LIME and MUSE (Moradi & Samwald, 2021).

| $N$ | LIME  | MUSE  | CIE   |
|-----|-------|-------|-------|
| 10  | 0.652 | 0.615 | 0.679 |
| 20  | 0.710 | 0.646 | 0.720 |
| 30  | 0.739 | 0.701 | 0.786 |
| 40  | 0.785 | 0.754 | 0.833 |
| 50  | 0.828 | 0.793 | 0.871 |

Table 3: The fidelity scores for different sizes of class-wise explanations on the outcomes of the LSTM classifier.

### 5 Conclusion

We presented the CIE explanation method that uses domain knowledge in combination with confident itemsets to extract semantic relationships between medical concepts and class labels within outcomes of black-box classifiers. Experiments on a disease-treatment information extraction task showed that CIE can approximate the black-boxes' behavior more accurately than the perturbation-based and decision set explanators. The interpretability experiments showed that CIE can generate more accurate approximations using smaller explanations than the other methods. The results demonstrate that confident itemsets, with the help of medical concepts, can accurately and efficiently identify decision boundaries of a black-box classifier in this type of BioNLP task. Revealing such information, CIE can help identify capabilities of complex models or biases in the problem and data.

Future work include extending CIE to other BioNLP tasks, and evaluating the ability of different black-box NLP models in capturing semantic relationships in a wide variety of BioNLP tasks.